\documentclass[lettersize,journal,twoside]{IEEEtran}
\usepackage{amsmath,amsfonts}

\usepackage[noend]{algpseudocode}
\usepackage{algorithm}
\usepackage{array}
\usepackage[caption=false,font=footnotesize,labelfont=rm,textfont=rm]{subfig}
\usepackage{textcomp}
\usepackage{stfloats}
\usepackage{url}
\usepackage{verbatim}
\usepackage{graphicx}
\usepackage{cite}
\usepackage{float}
\usepackage[T1]{fontenc}

\usepackage{amssymb}
\usepackage{booktabs}
\usepackage[table]{xcolor} \usepackage{pifont}
\usepackage{tabularx}
\usepackage{multirow}  
\usepackage{booktabs} 

\definecolor{customgreen}{RGB}{0, 153, 0}

\newcolumntype{Y}{>{\centering\arraybackslash}X}

\begin{document}
\title{Learning A Unified Risk Map for Autonomous Driving in Partially Observable Environments}

\author{Jie Jia$^{1}$, Yaofeng Su$^{1}$, Zeyu Bao$^{1}$, Yun Hong$^{1}$, Bingzhao Gao$^{2}$, Zhongxue Gan$^{1}$, and Wenchao Ding$^{1}$
\thanks{$^{1}$Jie Jia, Yaofeng Su, Zeyu Bao, Yun Hong, Zhongxue Gan and Wenchao Ding are with Fudan University, China (Corresponding author: Wenchao Ding). $^{2}$Bingzhao Gao is with Tongji University, China (e-mail:gaobz@tongji.edu.cn). Jie Jia and Yaofeng Su contributed equally to this work. E-mail: dingwenchao@fudan.edu.cn, 22210860105@m.fudan.edu.cn)}
\thanks{\textcopyright~ 2026 IEEE. Personal use of this material is permitted. 
Permission from IEEE must be obtained for all other uses, including 
reprinting/republishing this material for advertising or promotional 
purposes, creating new collective works, for resale or redistribution 
to servers or lists, or reuse of any copyrighted component of this 
work in other works. The final published version is available at: 
\protect\url{https://doi.org/10.1109/LRA.2026.3666393}.}}
% \thanks{This paper was recommended for publication by Editor Aniket Bera upon evaluation of the Associate Editor and Reviewers comments.}
% \thanks{This work was supported by the National Natural Science Foundation of China under Grant 62403142, and in part by the Science and Technology Commission of Shanghai Municipality under Grant 24511103100. (Corresponding author: Wenchao Ding.)}
% \texttt{dingwenchao@fudan.edu.cn}}
% \thanks{$^{1}$Jie Jia, Yaofeng Su, Zeyu Bao, Yun Hong, Zhongxue Gan and Wenchao Ding are with Fudan University, China \texttt{dingwenchao@fudan.edu.cn}}
% \thanks{$^{2}$Bingzhao Gao is with Tongji University, China \texttt{gaobz@tongji.edu.cn}}
% \thanks{$^{1}$Jie Jia and $^{1}$Yaofeng Su contributed equally to this work.}
% \thanks{Digital Object Identifier (DOI): see top of this page.}}
% \thanks{Manuscript received January XX, 2026. This work was supported by the National Natural Science Foundation of China under Grant 62403142, and in part by the Science and Technology Commission of Shanghai Municipality under Grant 24511103100. (Corresponding author: Wenchao Ding.)}
% \thanks{Jie Jia, Yaofeng Su, Zeyu Bao, Yun Hong, Zhongxue Gan and Wenchao Ding are with Fudan University, Shanghai 200433, China (e-mail: dingwenchao@fudan.edu.cn). Bingzhao Gao is with Tongji University, Shanghai 201804, China. Jie Jia and Yaofeng Su contributed equally to this work.}

% The paper headers
% \markboth{IEEE Robotics and Automation Letters. Preprint Version. Accepted February, 2026}%
\markboth{}%
{Jia \MakeLowercase{\textit{et al.}}: Learning A Unified Risk Map for Autonomous Driving in Partially Observable Environments}

%\IEEEpubid{0000--0000/00\$00.00~\copyright~2021 IEEE}
% Remember, if you use this you must call \IEEEpubidadjcol in the second
% column for its text to clear the IEEEpubid mark.

\maketitle

\begin{abstract}
Occlusion-aware prediction remains a critical challenge in autonomous driving due to the inherent uncertainty of unobserved regions. Existing approaches either overestimate risk based on reachable states or struggle to predict accurate trajectories under high occlusion uncertainty. To address these limitations, we propose a unified risk map modeling and learning framework for partially observable environments. Our method integrates traffic flow risk and collision risk through spatiotemporal modeling, enabling fine-grained assessment of occlusion-induced hazards. To address the scarcity of scenarios involving occluded interactions, we introduce a diffusion-based scenario generation framework that produces realistic yet adversarial scenarios. We integrate the modeling and learning of a unified risk map into a framework that supports risk-aware planning under partial observability. Experiments on the Waymo Open Motion Dataset show that our method significantly outperforms the state-of-the-art occlusion-aware baseline, improving minimum time-to-collision by 0.78 times and average time-to-collision by 1.67 times. The proposed framework offers a comprehensive and practical solution for risk-aware planning in partially observable environments.
\end{abstract}


\section{Introduction}

To address the challenges posed by visual occlusion and ensure the safe operation of autonomous driving systems, it is essential to assess potential occlusion risks beyond the field of view, thereby facilitating the formulation of safe driving strategies. Expert human drivers typically mitigate occlusion-related uncertainties by proactively decelerating to reduce potential risks. However, in real-world scenarios, interaction events with potential agents in occluded regions are relatively scarce. Consequently, directly relying on human driving data and employing mainstream imitation learning methods for driving strategy acquisition encounters significant bottlenecks. Under these circumstances, effectively anticipating and analyzing occlusion risks, as well as integrating them into the driving strategy planning process, emerges as a critical challenge in addressing occlusion uncertainty.

Existing occlusion-aware prediction methods fall into two main categories. Reachability-based approaches, such as those using Forward Reachable Sets (FRS)~\cite{orzechowski2018tackling, set2021}, evaluate all possible future states of hidden agents. While ensuring safety, they often lead to overly conservative planning by lacking data-driven traffic priors \cite{risk2019ral}. In contrast, learning-based methods~\cite{christianos2023planning,lange2024scene,itkina2022multi} predict trajectories or occupancy maps for hidden agents. However, they struggle to produce accurate predictions under the high uncertainty inherent in unobserved regions.
%\IEEEpubidadjcol

\begin{figure}[t]
  \centering 
  \includegraphics[width=0.35\textwidth, height=0.26\textheight]{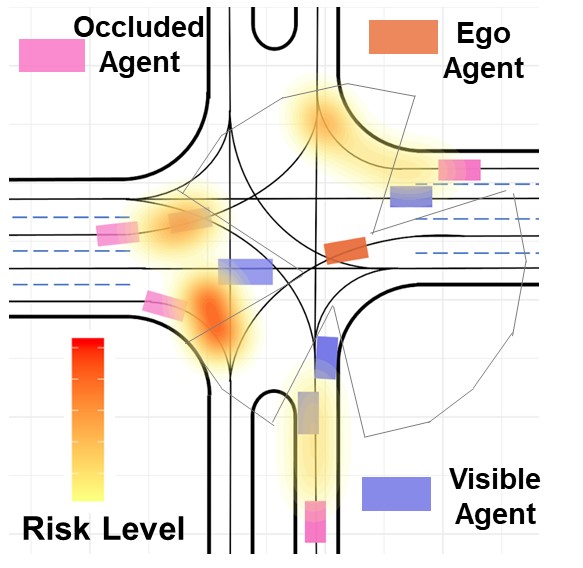}
  \caption{A unified spatiotemporal risk field integrates traffic flow and collision risks under partial observability. Flow risk is estimated via density from multimodal trajectory distributions, while collision risk is assessed by simulating the ego vehicle's trajectory to identify spatiotemporal hotspots.}

  \label{cover}
  \vspace{-0.5cm}
\end{figure}

To overcome these limitations, we propose a unified framework that rethinks how risk is modeled in partially observable environments. Our key insight is to construct a spatiotemporal risk field (Fig.\ref{cover}) that models underlying traffic flow density and potential collision hotspots. To address the data scarcity of critical occluded interactions, we introduce a diffusion-based generative model that produces realistic yet adversarial scenarios. This approach injects real-world traffic distributions into the learning process, mitigating the over-conservatism of reachability-based methods, while being more planning-friendly and stable than direct trajectory prediction. We integrate this risk field learning into a unified framework that supports risk-aware planning under partial observability.

We evaluate the effectiveness of the proposed framework through experiments on realistic occluded interaction scenarios from the Waymo Open Motion Dataset~\cite{ettinger2021large}. Qualitative results demonstrate that our approach accurately captures high-risk zones beyond the visible field and provides reliable risk distributions aligned with critical interaction points. Quantitative evaluations show that in challenging occlusion scenes, our method improves minimum time-to-collision by 0.78 times and average time-to-collision by 1.67 times compared to one state-of-the-art baseline. Our main contributions are summarized as follows:

\begin{itemize}
\item We propose a unified spatiotemporal risk field modeling framework in partially observable environments that combines traffic flow and  collision risks, enabling accurate and interpretable occlusion risk quantification.
\item We propose an automated method for generating occlusion scenarios that synthesizes realistic yet adversarial interactions to address the scarcity of rare but safety-critical occluded interaction data.
\item We integrate the modeling and learning of risk map to support risk-aware planning under partial observability. Experiments show that our method significantly outperforms the state-of-the-art occlusion-aware baselines.
\end{itemize}

\begin{figure*}
    \centering

    \includegraphics[width=0.836\textwidth, height=0.323\textheight]{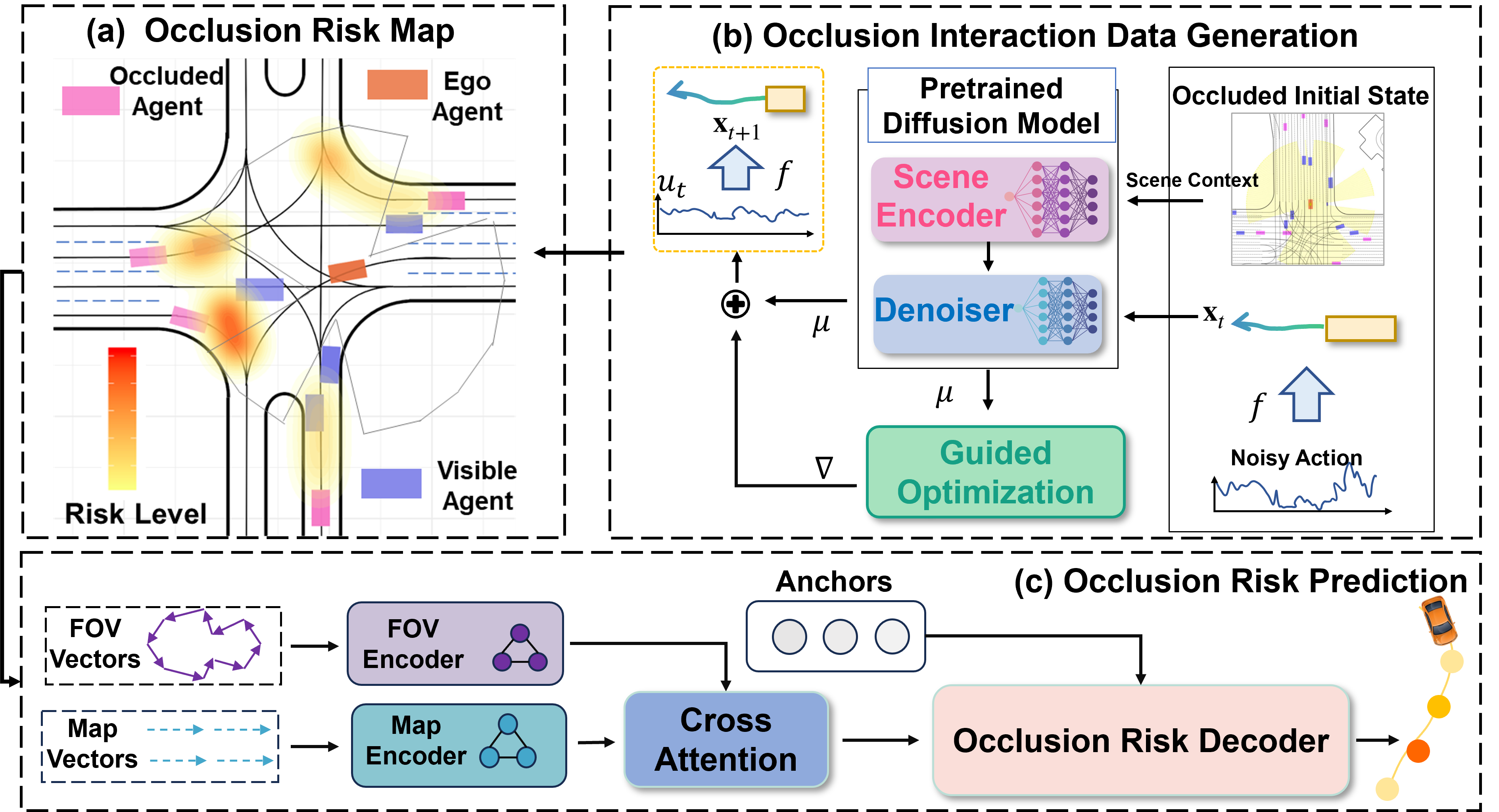}
    
    \caption{The framework of the proposed method: (a) a unified spatiotemporal risk field in partially observable environments; (b) an automated occlusion interaction data generator that samples potential occluded agent states and synthesizes realistic yet adversarial interactions using a pretrained diffusion model with guidance-based optimization; (c) a risk prediction model that encodes occlusion-aware environments and decodes probabilistic risk distributions.}
    \label{framework}
    \vspace{-0.3cm}
\end{figure*}

\section{Related Work}

\subsection{Occlusion Aware Prediction}

Occlusion-aware prediction research is primarily divided into analytical and data-driven approaches. Analytical methods~\cite{risk2020ICRA,risk2019ral,mcgill2019probabilistic,lee2017collision,wang2020generating} use formal techniques like reachability analysis to estimate future states of hidden agents. For instance, some works employ particle filtering~\cite{risk2019ral,risk2020ICRA} or incorporate vehicle semantics~\cite{risk2023ral} to refine risk estimation. Others utilize set-based approaches with Forward Reachable Sets (FRS) to ensure safety~\cite{orzechowski2018tackling, set2021}. However, these methods often overestimate risk, yielding conservative plans due to missing traffic priors.

The learning-based approaches predict trajectories or occupancy maps of occluded potential agents through occlusion inference \cite{afolabi2018people,itkina2022multi,christianos2023planning,lange2024scene,lei2025cooperriskdrivingriskquantification} for risk assessment. For instance, some works learn to predict occupancy grid maps for occluded regions based on the interactions of observed agents~\cite{afolabi2018people, itkina2022multi}. Christianos et al.~\cite{christianos2023planning} propose a two-stage training pipeline to predict future trajectories of inferred agents, along with a potential collision cost function for planning adjustment. Lange et al.~\cite{lange2024scene} introduce an attention-based single-stage method, Scene Informer, that jointly models both observed and occluded agents, providing trajectories for the former, and both occupancy probabilities and likely trajectories for the latter. Despite their data-driven nature that captures real traffic movement priors, these methods still face significant challenges in precisely predicting occluded trajectories due to the high uncertainty and unobservability of blind zones, which further impacts planning behaviors. Other works tackle partial observability through a Partially Observable Markov Decision Process (POMDP) framework~\cite{kurniawati2021partially}. For instance, Huang et al.~\cite{huang2024learningonlinebeliefprediction} propose an online belief update model to infer agents' intentions within an MCTS planner. While effective for POMDP-based planning, such specialized solutions are not always straightforward to integrate into more general motion planning systems. To overcome the limitations of previous occlusion-aware prediction works, this paper proposes a unified risk field modeling and prediction framework that improves the over-conservativeness of reachability-based methods through data-driven priors, while being more planning-friendly and reliable than trajectory prediction approaches under high uncertainty.

\subsection{Traffic Scenario Generation}

Traffic scenario generation, vital for autonomous driving development, involves initializing agent states and simulating their interactions. Early methods using replayed data or rule-based models~\cite{treiber2000congested,kesting2007general,zhang2024bayesian,waymo} often fail to reproduce complex, large-scale behaviors. Consequently, data-driven techniques have emerged to learn realistic priors from large datasets. Approaches include hierarchical imitation learning (BITS~\cite{xu2023bits}), socially controllable generation (SCBG~\cite{chang2023editing}), policy-search (MGAIL~\cite{igl2022symphony}), and diffusion-based synthesis (CTG~\cite{zhong2023guided}). However, these studies primarily focus on real-data distributions, with limited attention to simulating long-tail occluded interactions.

More recently, adversarial generators like STRIVE~\cite{rempe2022generating}, AdvDO~\cite{cao2022advdo}, KING~\cite{hanselmann2022king}, and CAT~\cite{zhang2023cat} have been developed to create safety-critical scenarios. Yet, these methods almost exclusively target visible-agent interactions, leaving occluded blind-zone simulations largely unaddressed. This motivates our work to develop an automated method for generating rare but critical occluded interaction scenarios.
\section{Method}

\subsection{Problem Statement}
This work addresses the problem of occlusion-aware reasoning for autonomous driving under partial observability. Formally, given the current observable environmental information $\mathcal{O}$, our goal is to find an optimal driving policy $\pi^*$ that also accounts for latent information $\mathcal{H}$ about hidden agents in occluded regions. The objective is to maximize safety and utility, conditioned on both observed and potential hidden information:
\begin{equation}
\pi^* = \arg\max_{\pi} \mathbb{E}_{\mathcal{H}}[ \mathcal{J}(\tau) | \mathcal{O}, \mathcal{H} ]
\end{equation}
where $\tau$ is the ego vehicle's future trajectory and $\mathcal{J}(\tau)$ represents the comprehensive cost function evaluating the safety, efficiency, and smoothness of the trajectory. Since $\mathcal{H}$ is unknown, the core challenge is to reason about this uncertainty. Our approach addresses this by first synthesizing a rich distribution of plausible yet adversarial scenarios to explicitly model the latent information $\mathcal{H}$, and from this, learning a unified spatiotemporal risk field that implicitly marginalizes over this uncertainty to guide the planner.
\subsection{Framework Overview}
To address the problem defined above, our framework, illustrated in Fig. 2, systematically tackles occlusion-aware reasoning through four interconnected components. We begin with \textbf{occlusion risk modeling}, constructing a dense, spatiotemporal risk field from fused traffic flow and collision risks. This model is trained on data from our \textbf{diffusion-based generator}, which synthesizes realistic yet adversarial scenarios. A lightweight \textbf{risk prediction network} then learns this risk representation for efficient real-time inference. Finally, a \textbf{risk-aware driving strategy} integrates the predicted risk into a downstream planner to ensure safe navigation. The following sections detail each component.

\subsection{Occlusion Risk Modeling}
To systematically model occlusion risks amidst perception uncertainty, we propose a continuous spatiotemporal risk field representation that captures both traffic flow dynamics and potential collision hotspots. Supported by our occlusion-aware data generator (Sec.~\ref{Method_Data_Generation}), this framework robustly models fine-grained risk distributions. It quantifies grid-level uncertainty by generating probabilistic traffic flow distributions from multimodal trajectories and identifies high-risk interactions by simulating collisions with the ego vehicle's planned path.

The process begins by preprocessing multimodal trajectory sets, expressed as ${T}_k = \{ \text{traj}_k^j \}_{j=1}^J$, where $J$ is the number of modes for the $k$-th agent. To focus on relevant hazards, we filter out stationary agents using a speed threshold $v_{\text{min}}$, yielding a set of active agents ${A}_{\text{active}}$. The map is then discretized into risk grids $\Omega$.

Our risk field comprises two components. First, \textbf{Flow Risk} is calculated based on the spatial density of predicted trajectories, indicating a higher risk where traffic is more likely to be present. It is quantified as:
\begin{equation}
    R_{\text{flow}}(n_1,n_2) = \sum_{T,J, {A}_{\text{active}}} {I}(\text{traj}_k^j(t), n_1,n_2) \cdot e^{-\lambda \cdot D}
\end{equation}
where ${I}(\cdot)$ is an indicator function for a trajectory point's presence within grid $\Omega(n_1,n_2)$, $D$ is the Euclidean distance to the grid center, and $\lambda$ is a spatial decay coefficient.

Second, \textbf{Collision Risk} quantifies the direct danger to the ego vehicle by detecting spatiotemporal overlaps. A collision event set ${C}_{\text{collision}}$ is first identified where the distance between the ego's trajectory and any predicted trajectory is less than a threshold $\delta$:
\begin{equation}
    {C}_{\text{collision}} =    \left\{ (t, x,y) \, \bigg| \, \| \text{traj}_{\text{ego}}(t) - \text{traj}_k^j(t) \|_2 < \delta \right\}.
\end{equation}
The collision risk field is then constructed from these events:
\begin{equation}
    R_{\text{collision}}(n_1,n_2) = \sum_{{C}_{\text{collision}}} {I}((t,x,y) , n_1,n_2) \cdot e^{-\lambda \cdot D}.
\end{equation}
Here, the variables ${I}(\cdot)$, $D$, and $\lambda$ have meanings analogous to those in the flow risk calculation but are applied to collision points.

Finally, the two risk components are linearly fused to form the total risk field, which serves as a dynamic safety map for the planner:
\begin{equation}
    R_{\text{total}}(n_1,n_2) = \alpha \cdot R_{\text{flow}}(n_1,n_2) + \beta \cdot R_{\text{collision}}(n_1,n_2),
\end{equation}
where $\alpha$ and $\beta$ are fusion weights. To enhance applicability, Gaussian filtering and normalization are applied to mitigate variations in scene scale and traffic density.

\subsection{Occlusion Interaction Data Generation} \label{Method_Data_Generation}

Owing to the high uncertainty and long-tail nature of occlusions, synthesizing corner-case interactions is a task for which prior generative models trained on real-scene data distributions~\cite{huang2024versatile, zhong2023guided} are ill-suited. We decompose this complex problem into two key subtasks: (1) estimating initial states of occluded agents, and (2) simulating their interaction strategies. Our diffusion-based framework first samples initial state distributions for potential agents and then employs a pretrained diffusion model to generate their trajectories, which are further optimized via a guidance function to enhance their adversarial nature.

\textbf{Initial State Generation for Occluded Agents.} 
To reasonably infer the initial states of potential vehicles in blind spots, we employ a probabilistic sampling-based method. Based on map topology and the ego vehicle's field of view, we sample start/end positions $[s_s, s_e]$ and speeds $[v_{min}, v_{max}]$ from a uniform distribution for potential agents within occluded lane segments. Each sample corresponds to a potential agent state, serving as a prior for the subsequent trajectory generation.

\textbf{Pretrained Diffusion Generative Model.} 
Benefiting from the initial state generation, our pretrained diffusion model generates occluded interaction trajectories. It predicts control sequences $u_t$ (acceleration $\dot{v}$ and yaw rate $\dot{\psi}$), which are then converted into state trajectories $x_t$ using a bicycle dynamics model. The diffusion model itself consists of a scene encoder and a denoiser, following the standard Denoising Diffusion Probabilistic Models (DDPM) framework~\cite{ho2020denoising}. The scene encoder utilizes a transformer-based architecture~\cite{attention,nayakanti2022wayformer} to process agent states and map data into a compact scene representation $\hat{c}$. The denoiser ${D}_\theta$ then reconstructs plausible trajectories by iteratively predicting controls at each step $k$, conditioned on $\hat{c}$ and noisy actions $\tilde{u}(k)$. The noise update at step $k$ follows established formulations~\cite{nichol2021improved}:
\begin{equation}
    \tilde{\mu}_k \gets \frac{\sqrt{\alpha_k}(1-\overline{\alpha}_{k-1})}{1-\bar{\alpha}_k}\tilde{u}_k + \frac{\sqrt{\bar{\alpha}_{k-1}}\beta_k}{1-\bar{\alpha}_k}\hat{a}
\end{equation}

\textbf{Guidance Function Optimization.}
While the pretrained diffusion model effectively captures the distribution of naturalistic driving behaviors, it inherently favors safe and nominal trajectories. However, training a robust risk map requires exposure to rare, safety-critical corner cases that are sparse in the original data distribution. To address this scarcity, we introduce a guidance function, inspired by recent works in controllable generation~\cite{dhariwal2021diffusion, jiang2023motiondiffuser}, to actively steer the generation process from nominal to adversarial modes. 

We model the occluded agent as an adversarial pursuer that seeks spatial conflict with the ego vehicle, subject to physical constraints. The optimization objective ${F}(u^p)$ is explicitly formulated to balance two competing goals: maximizing interaction risk (to provide valid supervision) and ensuring lane adherence (to maintain realism). Formally, the objective is defined as:
\begin{equation}
{F}(u^p) = \lambda_1 \min_t d_{\text{inter}}(x_t^p, x_t^{o}) + \lambda_2 \min_t d_{\text{road}}(x_t^p),
\label{eq:objective_function}
\end{equation}
where $x_t^p$ and $x_t^{o}$ are the states of the pursuer and other agents. The first term, $d_{\text{inter}}$, minimizes the distance at the closest point of approach to simulate near-misses or collisions. The second term, $d_{\text{road}}$, acts as a regularization constraint based on the Signed Distance Function (SDF), penalizing deviations from the road geometry. The weights $\lambda_1$ and $\lambda_2$ control the trade-off between adversarial intensity and physical plausibility.

During the reverse (denoising) process, this objective ${F}$ is maximized via gradient-based updates to the noise control sequence $\tilde{u}_k^p$ at each step $k$:
\begin{equation}
\tilde{u}_k^p \gets \tilde{u}_k^p + \lambda \sigma_k \nabla_{\tilde{u}_k^p} {F}({D}_\theta(\tilde{u}_k)),
\label{eq:update_rule}
\end{equation}
where $\lambda$ is a learning rate scaling factor and $\sigma_k$ is the noise standard deviation at step $k$. Crucially, by optimizing within the learned diffusion manifold rather than applying rigid heuristics, we ensure that adversarial behaviors remain grounded in naturalistic traffic distributions.

\subsection{Occlusion Risk Prediction}

To enable efficient and localized risk inference, our prediction model infers lane-anchored risk scores from vectorized environment representations. This is achieved using a transformer-based architecture~\cite{attention,nayakanti2022wayformer} where lane anchors serve as queries to decode scene features into occlusion risks.

\textbf{Occlusion Environment Encoding.} The model's input is a vectorized observation consisting of two sequences: the field of view (FOV) and the scene map, both aligned to the ego vehicle's coordinate frame. To capture perception in occluded environments, the visible region is encoded via a ray-tracing approach. The resulting FOV, represented as a set of rays (angle and distance), is processed by an MLP to produce the visibility encoding $\mathbf{f}_{\text{fov}}$. For map information, polylines with their attributes (e.g., position, lane type) are encoded via separate MLPs and aggregated through pooling to form a unified map representation $\mathbf{f}_{\text{map}}$. These visibility and map features are then fused via a cross-attention mechanism to produce a compact feature vector for downstream prediction:
\begin{equation}
\mathbf{f}_{\text{fused}} = \text{CrossAttention}(\mathbf{f}_{\text{fov}}, \mathbf{f}_{\text{map}})
\end{equation}

\textbf{Occlusion Risk Decoding.} Inspired by anchor-based prediction decoders popular in motion forecasting, such as in QCNet~\cite{zhou2023query}, we use lane anchors as queries to our risk prediction model. Lane anchors are key points selected along the lane space. Anchor sequences are mapped to feature space via an MLP to align with semantic elements. As shown in Fig.~\ref{framework}, anchor features are combined with temporal encodings and interact with the global occlusion features $\mathbf{f}_{\text{fused}}$ through attention to decode collision risk scores. The decoded features are processed through an MLP to generate multi-step risk predictions:
\begin{equation}
    y_{\text{risk}} = \text{TransformerDecoder}(\text{MLP}(\mathbf{A}), \mathbf{f}_{\text{fused}})
\end{equation}
where $\mathbf{A} \in \mathbb{R}^{B \times N_a \times D_a}$ is a batch of $N_a$ anchor embeddings with dimension $D_a$, and $y_{\text{risk}} \in \mathbb{R}^{B \times N_a}$ are the corresponding predicted risk scores along paths. The decoder uses a multi-layer Transformer~\cite{attention} to achieve spatiotemporal risk modeling. In inference, risk scores along planned trajectories are smoothed into continuous 2D risk fields via Gaussian filtering for refined risk assessment. The model is trained with a Mean Squared Error (MSE) loss between predicted and ground-truth risks.

\subsection{Driving Strategy}\label{驾驶策略}

Inspired by experienced drivers anticipate risks and slow down in occluded scenarios (e.g., intersections, alleys), we incorporate such foresight risks into general autonomous driving planning process. Specifically, our planner performs local trajectory generation along global references. Risk-aware planning is achieved via a composite cost function optimized through Quadratic Programming (QP):

\begin{equation}
\begin{split}
C_{\text{total}} =\ & w_1 \cdot C_{\text{smooth}} + w_2 \cdot C_{\text{reach}} \\
                    & + w_3 \cdot C_{\text{risk}} + w_4 \cdot C_{\text{collision}}
\end{split}
\end{equation}
Here, $w_i$ are weighting coefficients. 
$C_{\text{smooth}} = \sum_{i=1}^{n-1} (v_{i+1} - v_i)^2$ penalizes sudden accelerations; 
$C_{\text{reach}} = \sum_{i=1}^n (d_i - d_{\text{desired}})^2$ encourages the trajectory to reach the target; 
$C_{\text{risk}} = \sum_{i=1}^n R_i \cdot v_i^2$ accounts for predicted occlusion risk, discouraging high speeds in risky regions; 
and $C_{\text{collision}} = \sum_{t=1}^T \exp(-d_t)$ penalizes proximity to visible obstacles. By minimizing this cost, the planner can generate expert-level risk-aware trajectories in occluded environments.
\begin{figure*}[ht]
    \centering
    \subfloat[Interaction Trajectory\label{fig:scene1c_}]{
        \includegraphics[width=0.2\textwidth,height=0.2\textwidth]{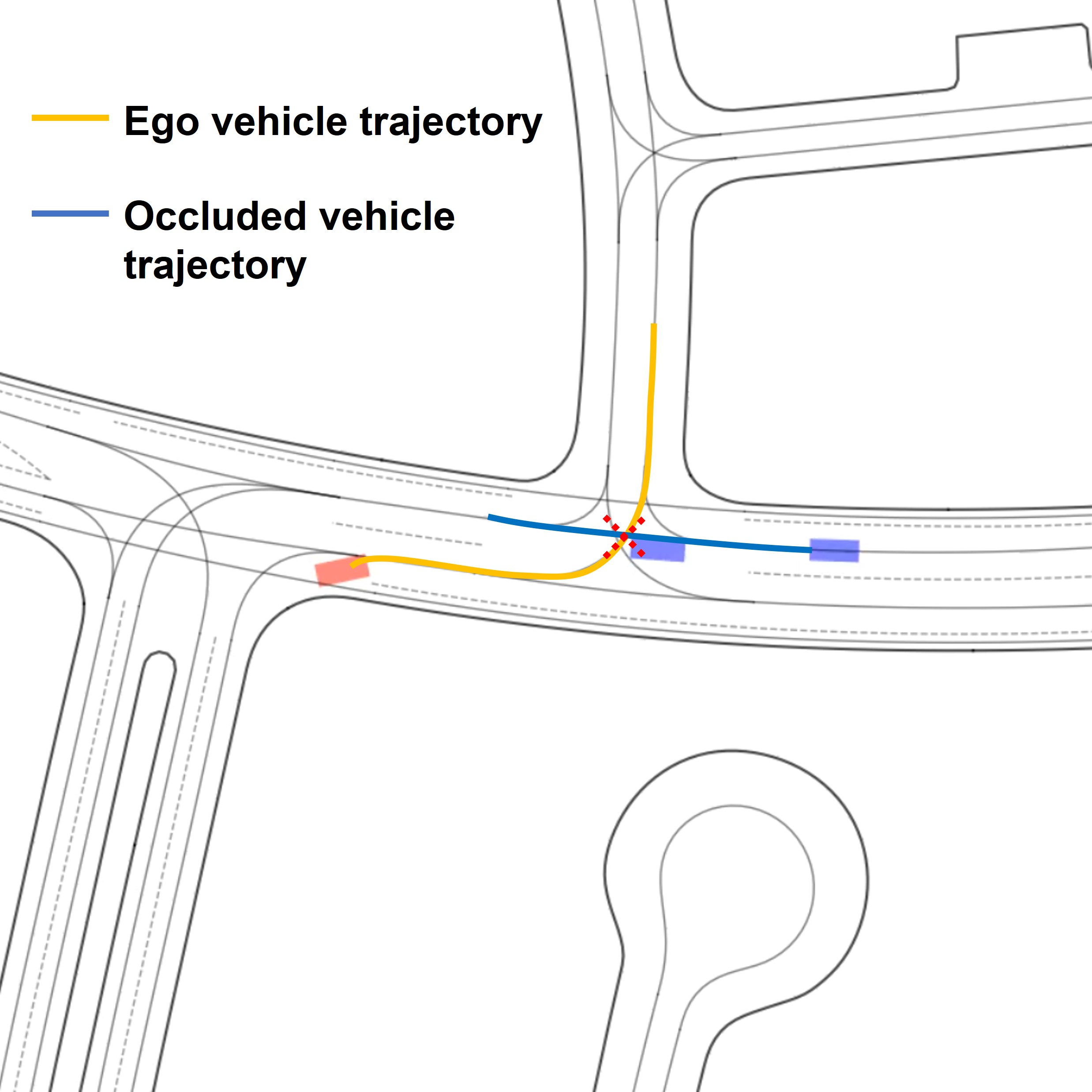}}
    \hfil
    \subfloat[Occluded by Dynamic Obstacle\label{fig:scene1a_}]{
        \includegraphics[width=0.2\textwidth,height=0.2\textwidth]{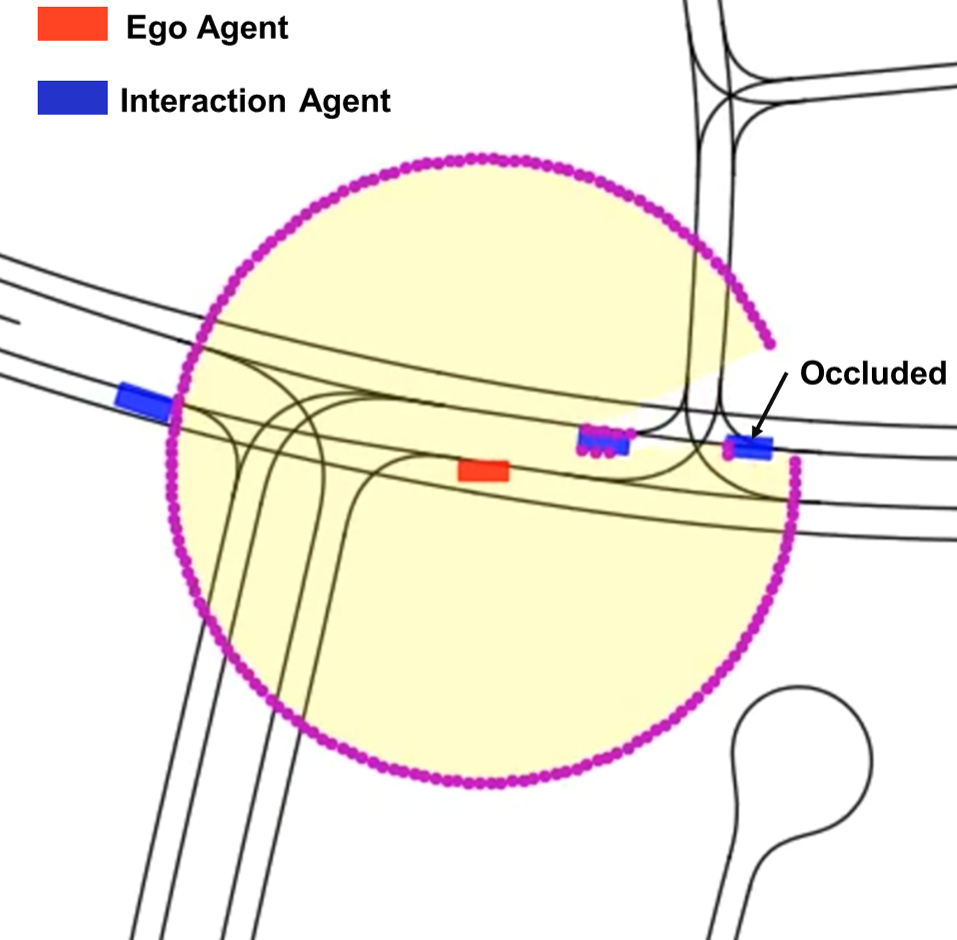}}
    \hfil
    \subfloat[Occluded Vehicle observed\label{fig:scene1b_}]{
        \includegraphics[width=0.2\textwidth,height=0.2\textwidth]{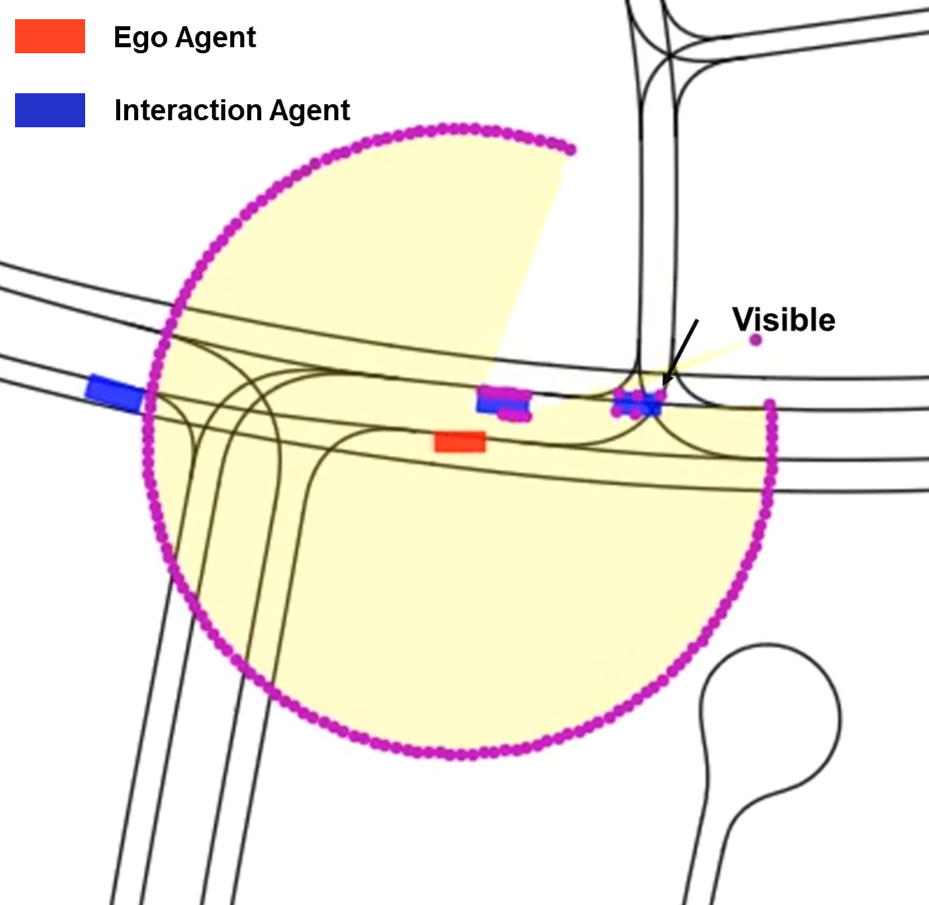}}
    \hfil
    \subfloat[Velocity-Distance Profile\label{velocity_distance_profile}]{
        \includegraphics[width=0.25\textwidth,height=0.2\textwidth]{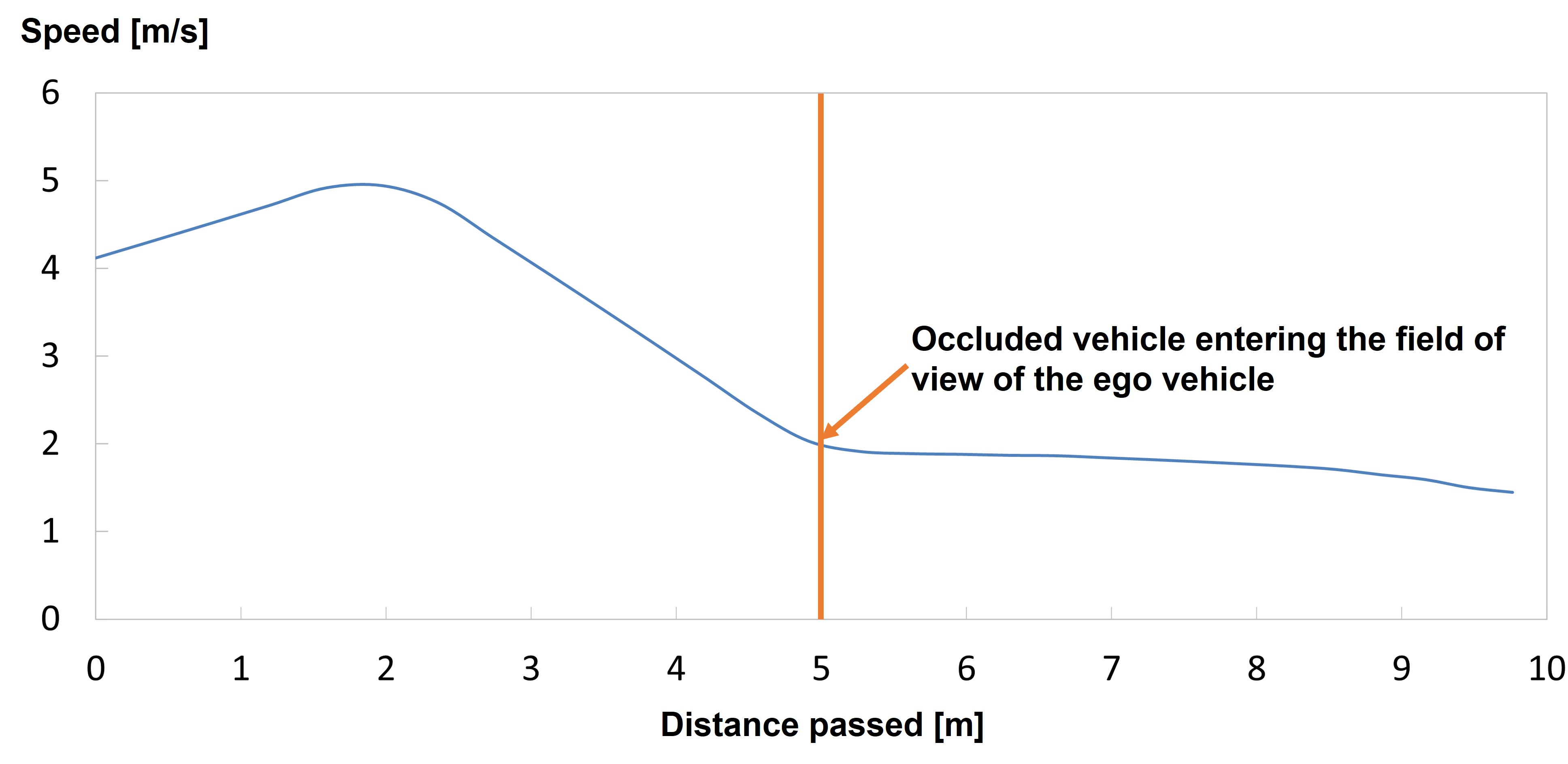}}
    
    \caption{Qualitative Analysis of Planning Performance.}
    \label{planning_performance_test_scenario}
    \vspace{-0.5cm}
\end{figure*}

\section{Experiments}

\subsection{Experimental Setup}

Experiments are conducted on the Waymo Open Motion Dataset (WOMD)\cite{ettinger2021large}, a large-scale open-source dataset containing recorded object trajectories and corresponding scene maps across diverse real-world driving scenarios. Each scenario from WOMD has a duration of 9 seconds at 10 Hz; we use the initial 8 seconds for our experiments to ensure data consistency across scenarios. We chose WOMD because its high-quality off-board perception labels provide an excellent foundation for learning risk models from diverse, real-world occluded scenarios. While it may contain fewer hand-crafted, complex adversarial cases than some simulators, we believe its scale and realism make it a superior choice for developing generalizable models.

To evaluate the planning improvement enabled by risk prediction in long-tail occlusion scenarios, we select 1,000 training scenes with potential perception uncertainty beyond the field of view from the WOMD training set for risk field modeling and prediction training, and 100 validation scenes from the WOMD validation set. The validation scenes represent real-world occlusion cases where occluded agents interact with the ego vehicle. In the planning evaluation, we follow a common protocol where the ego vehicle's planner controls its velocity profile along a fixed reference path from the dataset. This open-loop setting allows for a fair and direct comparison of how different risk assessment methods influence planning, especially when benchmarking against prediction-focused baselines.

\subsection{Implementation Details}

\textit{Risk Field Modeling:} The risk field is constructed based on multimodal trajectory prediction results, supported by our occlusion scenario generation method (Section~\ref{Method_Data_Generation}). For each scene, we include the initial states of sampled occluded vehicles and the multimodal trajectory distributions of all traffic participants. Using these inputs, the proposed method computes both traffic flow and potential collision risks across the scene, generating a normalized risk grid map with 0.5\,m resolution and values scaled to [0,1]. All experiments are conducted on a workstation equipped with an Intel Xeon Gold 6133 CPU and an NVIDIA RTX 4090 GPU.

\textit{Risk Prediction Network Training:} The training set is constructed by uniformly sampling 20 anchor points along the ego vehicle trajectories from the WOMD~\cite{ettinger2021large} dataset and caching the corresponding risk values derived from the risk modeling at these anchors. The network is trained with a learning rate of $1 \times 10^{-5}$ and a batch size of 4, with the model checkpoint selected at epoch 100. The network's anchor point risk outputs are interpolated to reconstruct complete inference risk fields for planning verification.

\textit{Computational Performance:} On an NVIDIA RTX 4090 GPU, the complete model achieves an average inference latency of 6.67 ms (150 FPS), satisfying real-time requirements. Component-wise analysis reveals minimal overhead: the ray-tracing-based FOV encoding requires {1.56 ms} (23.4\% of total latency), and the Transformer-based risk decoder accounts for {3.45 ms} (51.7\% of total latency).

\subsection{Comparative Methods}

To validate the effectiveness of our proposed risk modeling and prediction network, this study compares the planned trajectories generated by two ablation variants and two state-of-the-art baselines with our approach under identical test scenarios:

\textit{Ablation Variants (NOAP \& O-Risk)}: Two variants are included to isolate key contributions. 
{Non-Occlusion-Aware Planning (NOAP)} performs trajectory planning without considering occlusion risks. This method uses the same planning framework as ours but does not account for occlusion risks during planning, serving to analyze the impact of risk awareness. 
{Original-Risk (O-Risk)} validates the contribution of the scenario generation framework. It utilizes the proposed risk prediction network but is trained solely on raw WOMD data without the diffusion-generated scenarios. Comparing this variant with our full method isolates the gain attributed to the augmented adversarial data.

\textit{Baseline1 (Occlusion-Prediction-Based Planning, OPBP)}: Trajectory planning based on the state-of-the-art occlusion trajectory prediction method Scene Informer\cite{lange2024scene}. This method considers potential risks in occluded areas by reasoning about blind zones and outputting predicted trajectories of occluded agents, then evaluates these predicted trajectories using the collision cost term $C_{\text{collision}}$.

\textit{Baseline2 (Reachability-based Planning, SRQ-P):} To ensure a diverse comparison, we include a representative method from the reachability-based paradigm, proposed by Park et al. \cite{risk2023ral}. This approach uses Simplified Reachability Quantification to efficiently compute the risk posed by potential phantom agents in occluded areas. The quantified risk is then translated into a dynamic speed limit, which serves as a hard constraint for re-planning on the ego vehicle's original path.

Our proposed method builds upon the architecture of NOAP but integrates the unified risk map learned from the augmented data (unlike O-Risk) into the trajectory planning process to enhance the rationality and safety of planning results.

\subsection{Evaluation Metrics}

TTC\textsubscript{min} (Minimum Time-to-Collision): The average minimum time (in seconds) to a collision between the ego vehicle and any other traffic participant across all timesteps. This metric evaluates extreme risk; smaller values indicate higher collision risks.

TTC\textsubscript{avg} (Average Time-to-Collision): The average time (in seconds) to a collision across all timesteps and all agent pairs. This metric measures the average risk throughout the entire interaction; larger values indicate higher overall safety.

Risk Score: Evaluates the risk of each trajectory under Ground Truth risk field labels using the formula:
\begin{equation}
\sum_{i=0}^{N-1} \text{risk}(t_i) \times \text{velocity}(t_i) \times \Delta t
\end{equation}

The risk values are obtained through risk field modeling of flow risk and collision risk on grid maps. Smaller values indicate lower probabilities of collisions or close interactions between the planned trajectory and other traffic participants.

Critical Moments: Evaluate the average number of time frames when the ego vehicle has less than 3 seconds before colliding with others.

\begin{figure*}[t]
    \centering
    % 注意：\subfloat[子标题\label{子标签}]{图片命令}
    \subfloat[Scenario a: unprotected left-turn occluded intersection\label{fig:scene1a}]{
        \includegraphics[width=0.225\textwidth,height=0.225\textwidth]{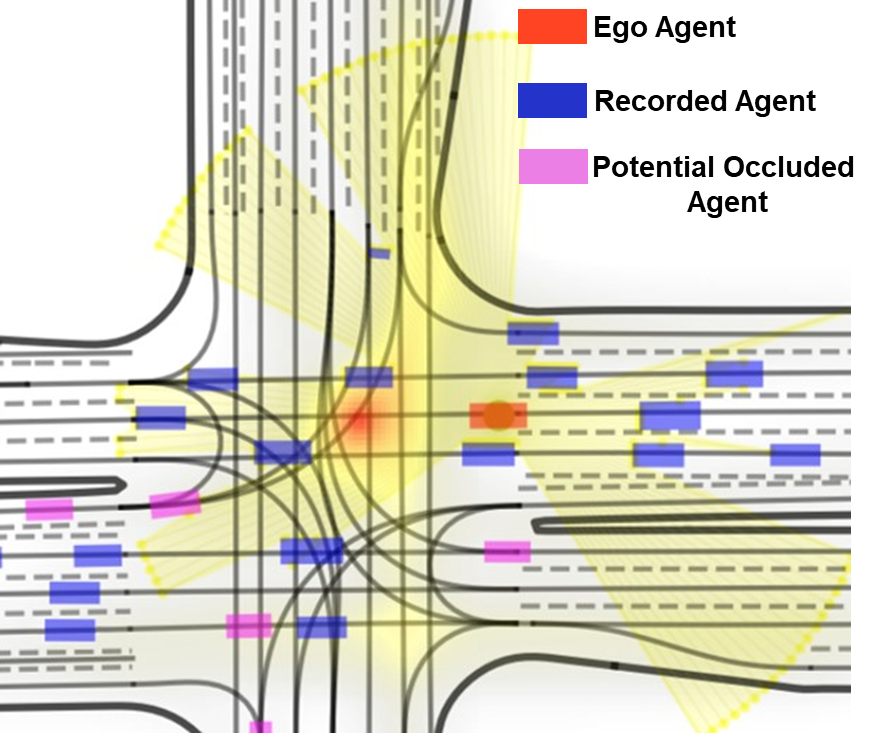}}
    \hfil % 增加间距
    \subfloat[Scenario b: an occluded T-intersection\label{fig:scene1b}]{
        \includegraphics[width=0.225\textwidth,height=0.225\textwidth]{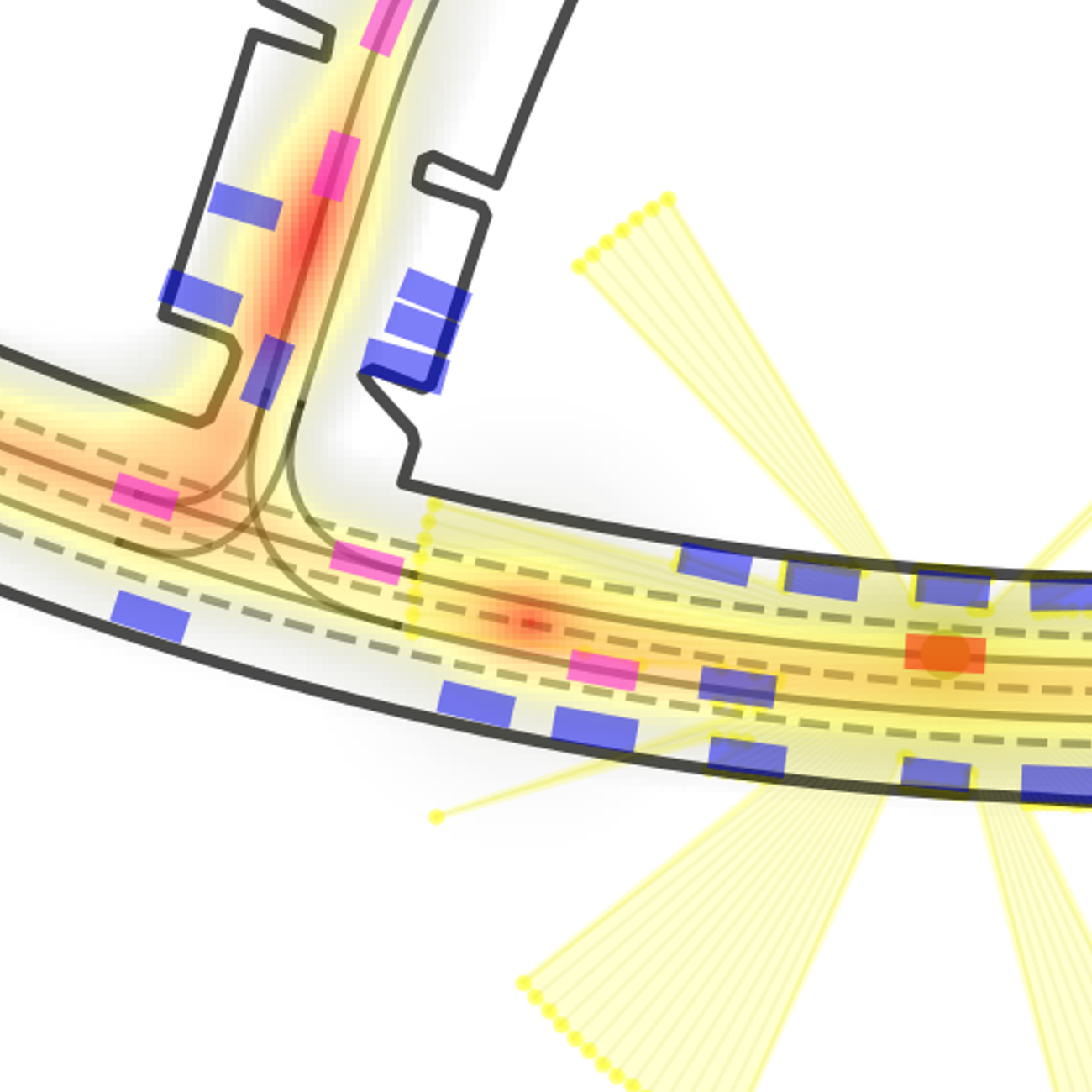}}
    \hfil
    \subfloat[Scenario c: a complex occluded narrow road\label{fig:scene1c}]{
        \includegraphics[width=0.225\textwidth,height=0.225\textwidth]{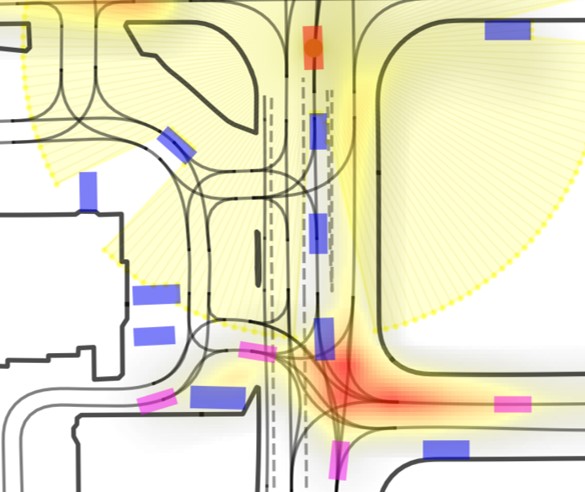}}
    \hfil
    \subfloat[Scenario d: a occluded intersection\label{fig:scene1d}]{
        \includegraphics[width=0.225\textwidth,height=0.225\textwidth]{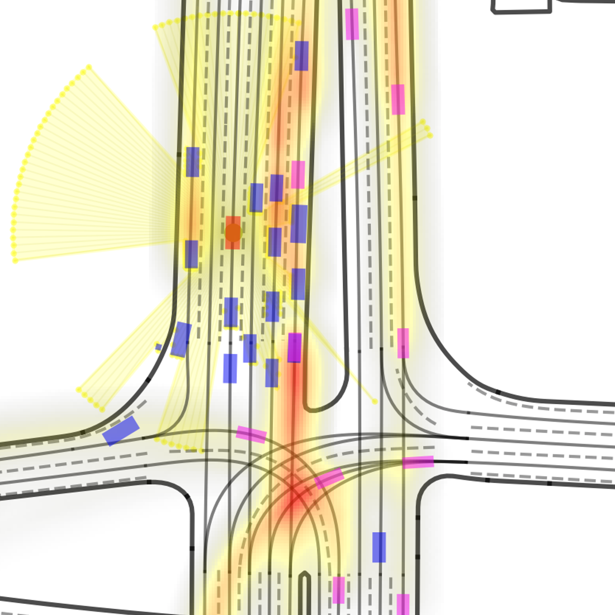}}
    
    \caption{Results of Risk Field Modeling. Red vehicles denote the ego vehicle, blue vehicles represent recorded agents from the WOMD, and pink vehicles are potential occluded agents sampled by our generation framework. The light yellow area shows the ego vehicle's visible range. The modeled risks are visualized using a white-to-yellow-to-red gradient heatmap, where red indicates the highest risk.}
    \label{fig:risk_modeling_effects}
    \vspace{-0.5cm}
\end{figure*}

\subsection{Effect Analysis}

\textbf{Quantitative Analysis of Planning Performance.}
To validate our risk prediction model's effectiveness in occluded interaction scenarios, we compare its performance against several baselines on identical scenarios, where quantitative results demonstrate its significant performance improvements.

\begin{table}[h]
    \centering
    \caption{Comparison of Planning Performance}
    \renewcommand{\arraystretch}{1.5}
    
    \begin{tabular}{>{\centering\arraybackslash}p{1.9cm} 
    >{\centering\arraybackslash}p{1.0cm} 
    >{\centering\arraybackslash}p{1.0cm} 
    >{\centering\arraybackslash}p{1.0cm} 
    >{\centering\arraybackslash}p{1.2cm}
    }
        \hline
        {\textbf{Method}} & 
        \text{TTC\textsubscript{min}} (s)$\uparrow$ & 
        \text{TTC\textsubscript{avg}} (s)$\uparrow$ & 
        Risk Score$\downarrow$ & 
        Critical Moments$\downarrow$ \\
        \hline
        NOAP & 3.59 & 7.59 & 1.38 &16.29 \\
        O-Risk & {4.91} &{11.57} &  0.71 & {10.02}\\
        \hline
        SRQ-P \cite{risk2023ral} & {4.37} & {9.09} & {0.30} & {10.64} \\
        OPBP\cite{lange2024scene} & 4.32 & 13.16 & 1.24 &14.35 \\
        \hline
        \textbf{Our*} & {7.72} &{35.14} &  0.68 & {5.41}\\
        \hline
    \end{tabular}
    \label{Quantitative Analysis of Planning Performance}
\end{table}

% \textit{Comparison with NOAP:} As an ablation, NOAP performs the worst across all metrics, demonstrating that ignoring occlusion risk leads to significant safety challenges in uncertain interactions. In contrast, our method's ability to effectively predict and integrate risk information results in a substantial improvement in planning safety.

\textit{Ablation Analysis (NOAP \& O-Risk):} As an ablation, NOAP performs the worst across all metrics (e.g., TTC\textsubscript{min} of 3.59s), demonstrating that ignoring occlusion risk leads to significant safety challenges. While O-Risk improves safety (4.91s), it still lags behind the full method (7.72s). This gap indicates that the deterministic, single-trajectory nature of raw logs is insufficient for modeling probabilistic risk fields and lacks critical corner cases. In contrast, our augmented data successfully bridges these gaps, yielding the most substantial safety improvements.

\textit{Comparison with OPBP:} The OPBP baseline, which improves safety over NOAP by predicting occluded agent trajectories, is still significantly outperformed by our method. Specifically, our approach yields improvements of {0.78 times} in minimum TTC and {1.67 times} in average TTC over OPBP, along with a {62.3\%} reduction in critical moments. This significant improvement in safety metrics demonstrates that our unified risk field provides a more comprehensive and stable representation of risk under high occlusion uncertainty compared to relying on explicit trajectory predictions, thereby more effectively reducing high-risk interactions.

\textit{Comparison with SRQ-P:} The reachability-based method, SRQ-P, demonstrates strong safety performance, particularly when compared to the NOAP and OPBP baselines. However, our proposed method achieves a {76.7\%} higher minimum TTC and a {287\%} higher average TTC, indicating significantly larger safety margins during interactions. The lower Risk Score of SRQ-P may be attributed to its conservative velocity planning, as the risk score formulation heavily penalizes high speeds in risky areas. In contrast, our unified risk map provides a more nuanced risk assessment as evidenced by the superior TTC and Critical Moments metrics.

\textbf{Qualitative Analysis of Planning Performance.}
As shown in Fig.~\ref{planning_performance_test_scenario}, our experimental scenario features two consecutive vehicles passing through an intersection, where the leading vehicle dynamically occludes the following one. This prevents the ego vehicle from fully perceiving the approaching traffic until the lead vehicle has passed. The global route map reveals that the ego's planned left-turn trajectory creates a potential collision risk with this occluded vehicle, highlighting how a failure to predict risks in advance can lead to collisions.

Fig.~\ref{velocity_distance_profile} presents the velocity-distance profile from our method, demonstrating its effectiveness in risk-aware planning. The velocity profile shows that our approach proactively decelerates based on predicted risks \textit{before} the occluded vehicle becomes visually detectable. This strategy significantly reduces occlusion uncertainty and provides sufficient time for safe interaction, ensuring passage with minimal risk exposure. The method's strong performance in such high-safety-guarantee scenarios validates its effectiveness in risk-predictive driving strategy planning.

\textbf{Occlusion Scenario Generation Analysis.} To validate our diffusion-based framework for generating realistic yet adversarial occlusion scenarios, we conducted a dedicated quantitative analysis. The experiment was performed on the same 100 validation scenarios from WOMD~\cite{ettinger2021large} used in our main planning evaluation. Our diffusion model was pretrained on the WOMD training set using an MSE loss consistent with standard DDPMs~\cite{ho2020denoising}, a learning rate of 2e-4, a batch size of 6, and for 16 epochs. For comparison, we benchmarked against two baselines: \textbf{Log-Replay}, which uses the original dataset trajectories, and a \textbf{Rule-based} method~\cite{waymo} that applies a constant velocity model to the same initial agent states sampled by our approach. We evaluated the generated scenarios based on their adversarial quality (Average Time-to-Collision, Interaction Agents Number) and realism (OnRoad Rate, OffRoad Distance).

\begin{table}[h]
    \centering
    \caption{Comparison of Occlusion Scenario Generation}
    \renewcommand{\arraystretch}{1.5}
    
\begin{tabular}{>{\centering\arraybackslash}p{1.9cm} 
    >{\centering\arraybackslash}p{1.0cm} 
    >{\centering\arraybackslash}p{1.0cm} 
    >{\centering\arraybackslash}p{1.0cm} 
    >{\centering\arraybackslash}p{1.2cm}
    }
        \hline
        \textbf{Method} & 
        TTC (s)$\downarrow$ & 
        OnRoad Rate (\%)$\uparrow$ & 
        OffRoad Dist (m)$\downarrow$ & 
        Int. Agents Num \\
        \hline
        Log-Replay          & 70.36          & 100.00         & 0.00           & 24.99 \\
        Rule-based~\cite{waymo} & 41.68          & 81.38          & 10.74          & 37.42 \\
        \textbf{Our*}       & {24.52} & {92.44} & {4.66}  & {37.42} \\
        \hline
    \end{tabular}
    \label{tab:scenario_generation_comparison}
\end{table}

As shown in Table~\ref{tab:scenario_generation_comparison}, our method achieves a superior balance of realism and adversarial quality. Compared to the Rule-based approach, it generates far more challenging scenarios (TTC reduced by {41.2\%}) while maintaining higher realism (OnRoad Rate up {13.6\%}, OffRoad Distance down {56.6\%}). Moreover, our initial state generation strategy increases the number of interacting agents by {49.7\%} over the original log, confirming its ability to synthesize complex, safety-critical scenarios that are both plausible and challenging, providing a strong foundation for our risk modeling.

\textbf{Risk Map Modeling Analysis.}
Fig.~\ref{fig:risk_modeling_effects} validates our risk modeling effectiveness across four diverse scenarios. \textbf{In (a)}, where the view is occluded at an intersection with shared lanes, our model correctly identifies primary risks from potential left-turning agents in the opposite direction. \textbf{In (b)}, at a dense T-intersection with blocked visibility, it captures hazards from vehicles that could be traveling in blind zones or suddenly entering from side roads. \textbf{For (c)}, it robustly models multi-directional threats on a narrow, highly occluded road. \textbf{Finally, in (d)}, it demonstrates long-horizon foresight by successfully predicting risks at potential lane convergence zones, even when the ego vehicle is still far from a blocked intersection. Across all cases, our approach accurately simulates occluded agents and identifies high-risk areas beyond the visible field, confirming its practical effectiveness and generalization.

\begin{figure}[h] % 注意这里你是单栏 figure，如果太宽可以用 figure*
    \centering
    \subfloat[Scenario a: unprotected left turn at occluded intersection\label{fig:pred_scene1a}]{
        \includegraphics[width=0.20\textwidth,height=0.20\textwidth]{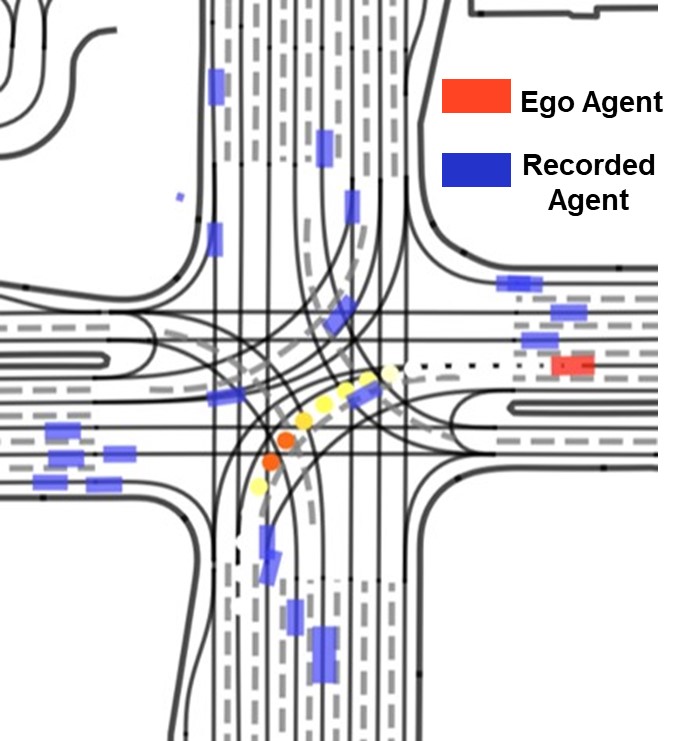}}
    \hfil
    \subfloat[Scenario b: occluded T-intersection\label{fig:pred_scene1b}]{
        \includegraphics[width=0.20\textwidth,height=0.20\textwidth]{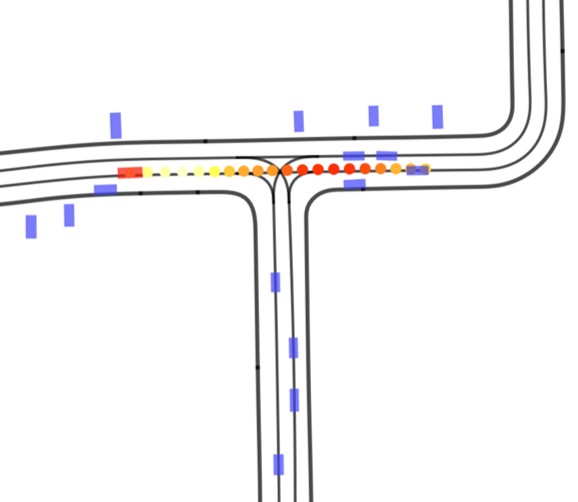}}
    
    % 如果你想换行，可以加个空行或者 \\
    \vspace{0.2cm} % 增加垂直间距
    
    \subfloat[Scenario c: occluded consecutive intersections\label{fig:pred_scene1c}]{
        \includegraphics[width=0.20\textwidth,height=0.20\textwidth]{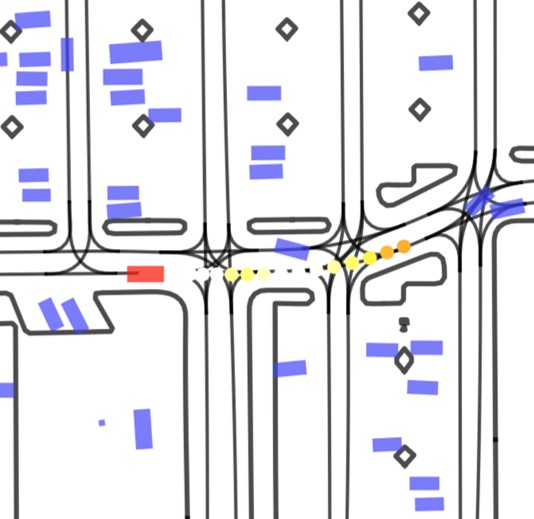}}
    \hfil
    \subfloat[Scenario d: occluded crossroad\label{fig:pred_scene1d}]{
        \includegraphics[width=0.20\textwidth,height=0.20\textwidth]{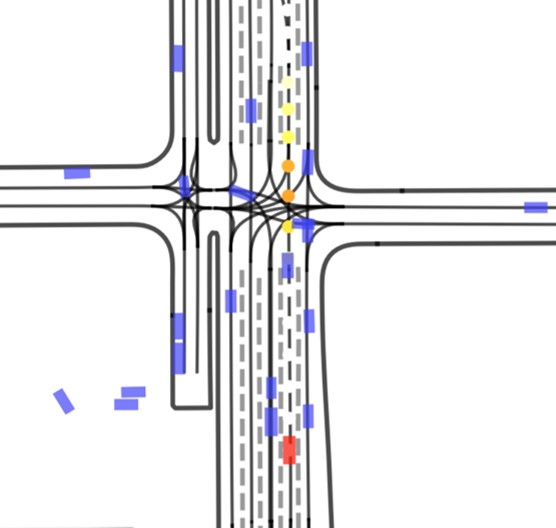}}
        
    \caption{Results of Risk Prediction. Red denotes the ego vehicle, blue indicates recorded agents from WOMD. The white-yellow-red gradient points represent lane anchors with predicted risk at equal future time intervals. The heatmap intensity reflects the predicted risk level, with red signifying the highest risk, allowing for intuitive visualization of how the model anticipates hazards along the ego-vehicle's path.}
    \label{risk_prediction_model_effectiveness}
\end{figure}

\textbf{Risk Prediction Model Analysis.}
The performance of our risk prediction model is demonstrated in Fig.~\ref{risk_prediction_model_effectiveness}. \textbf{In Scenario (a)}, during a complex left turn, the model accurately reflects real dynamics by capturing low initial risk that peaks precisely at conflict points with oncoming traffic. \textbf{For (b)}, it shows strong warning capability by providing early risk signals for possible collisions with right-turning vehicles emerging from a side road. \textbf{In (c)}, while navigating two adjacent intersections, the model successfully identifies distinct high-risk zones at both locations. \textbf{Finally, in (d)}, where the ego vehicle approaches a dead-end within a multi-intersection layout, it effectively detects high risk regions at potential interaction points. These results confirm the model's ability to accurately localize risk areas and dynamically adjust risk intensity over time, offering reliable support for safe decision-making.

\section{CONCLUSIONS}
This paper introduced a unified framework for risk-aware planning in occluded environments by integrating a novel spatiotemporal risk model with an adversarial, diffusion-based scenario generator. Our lightweight prediction network enables efficient, lane-anchored risk inference, leading to significant safety improvements on the Waymo dataset, including substantially increased time-to-collision metrics. Future work will focus on enhancing traffic priors via improved diffusion-based sampling.

% \section*{Acknowledgments}
% This should be a simple paragraph before the References to thank those individuals and institutions who have supported your work on this article.

% {\appendix[Proof of the Zonklar Equations]
% Use $\backslash${\tt{appendix}} if you have a single appendix:
% Do not use $\backslash${\tt{section}} anymore after $\backslash${\tt{appendix}}, only $\backslash${\tt{section*}}.
% If you have multiple appendixes use $\backslash${\tt{appendices}} then use $\backslash${\tt{section}} to start each appendix.
% You must declare a $\backslash${\tt{section}} before using any $\backslash${\tt{subsection}} or using $\backslash${\tt{label}} ($\backslash${\tt{appendices}} by itself
%  starts a section numbered zero.)}

% %{\appendices
% %\section*{Proof of the First Zonklar Equation}
% %Appendix one text goes here.
% % You can choose not to have a title for an appendix if you want by leaving the argument blank
% %\section*{Proof of the Second Zonklar Equation}
% %Appendix two text goes here.}

% \section{References Section}
% You can use a bibliography generated by BibTeX as a .bbl file.
%  BibTeX documentation can be easily obtained at:
%  http://mirror.ctan.org/biblio/bibtex/contrib/doc/
%  The IEEEtran BibTeX style support page is:
%  http://www.michaelshell.org/tex/ieeetran/bibtex/
 
%  % argument is your BibTeX string definitions and bibliography database(s)
% %\bibliography{IEEEabrv,../bib/paper}
% %

\bibliographystyle{IEEEtran} % 设置参考文献风格为 IEEE 标准
\bibliography{ref}  % 调用数据库。ref 对应你的 ref.bib 文件名

% \newpage

% \section{Biography Section}
% If you have an EPS/PDF photo (graphicx package needed), extra braces are
%  needed around the contents of the optional argument to biography to prevent
%  the LaTeX parser from getting confused when it sees the complicated
%  $\backslash${\tt{includegraphics}} command within an optional argument. (You can create
%  your own custom macro containing the $\backslash${\tt{includegraphics}} command to make things
%  simpler here.)
 
% \vspace{11pt}

% \bf{If you include a photo:}\vspace{-33pt}
% \begin{IEEEbiography}[{\includegraphics[width=1in,height=1.25in,clip,keepaspectratio]{fig1}}]{Michael Shell}
% Use $\backslash${\tt{begin\{IEEEbiography\}}} and then for the 1st argument use $\backslash${\tt{includegraphics}} to declare and link the author photo.
% Use the author name as the 3rd argument followed by the biography text.
% \end{IEEEbiography}

% \vspace{11pt}

% \bf{If you will not include a photo:}\vspace{-33pt}
% \begin{IEEEbiographynophoto}{John Doe}
% Use $\backslash${\tt{begin\{IEEEbiographynophoto\}}} and the author name as the argument followed by the biography text.
% \end{IEEEbiographynophoto}

% \vfill

\end{document}